\documentclass[conference]{IEEEtran}
\usepackage[american]{babel}
\usepackage{graphicx}
\usepackage{hyperref}
\usepackage{url}
\usepackage{orcidlink}
\title{Reasoning, Code, or Both? How Large Language Models Handle Variations in Math Questions}
\author{\IEEEauthorblockN{Matthew Kutakh \orcidlink{0009-0009-7193-8294}}}

\IEEEoverridecommandlockouts
\IEEEpubid{\copyright~2026 Matthew Kutakh}

\begin{document}
\maketitle

\begin{abstract}
Large Language Models (LLMs) achieve impressive accuracy on mathematical reasoning benchmarks, yet their performance drops when problems are modified with simple changes like different names or numbers. Code execution methods, which let models generate and run Python code instead of reasoning in natural language, have been proposed as a solution, but their effect on reasoning robustness (the ability to maintain accuracy across problem variations) has not been systematically tested. This study evaluates three approaches on 1,000 problems from the GSM-Symbolic dataset: pure reasoning using chain-of-thought (CoT) prompting, single-shot code execution using Program-Aided Language models (PAL), and iterative code execution using Step-by-Step Coding (SBSC). All three were run on paired original and modified problems using Claude Haiku 4.5. CoT was the most robust method, with an accuracy drop of 1.3 percentage points and 1.8\% of problems breaking under perturbation. PAL was the least robust at 1.7 percentage points and 3.1\% broke, with SBSC falling in between. Although these differences were not statistically significant ($p = .096$), the directional trend was consistent across all measures, suggesting that code execution, whether single-shot or iterative, does not improve reasoning robustness on grade-school-level problem variations.
\end{abstract}


\section{Introduction}
Large Language Models (LLMs) have shown an extraordinary performance, with top-of-the-line models achieving a gold medal level on International Math Olympiad problem sets \cite{huang2025}. Despite that, their performance drops dramatically with simple modifications and unnecessary additions to the simple math questions \cite{mirzadeh2025}, raising the fundamental question of the reasoning robustness of these models. Enter iterative code execution, a promising approach that transforms unreliable language into executable code \cite{gao2023,singh2025}, but one whose true effectiveness remains systematically unexplored.

\begin{figure*}[!t]
    \centering
    \includegraphics[width=0.7\textwidth]{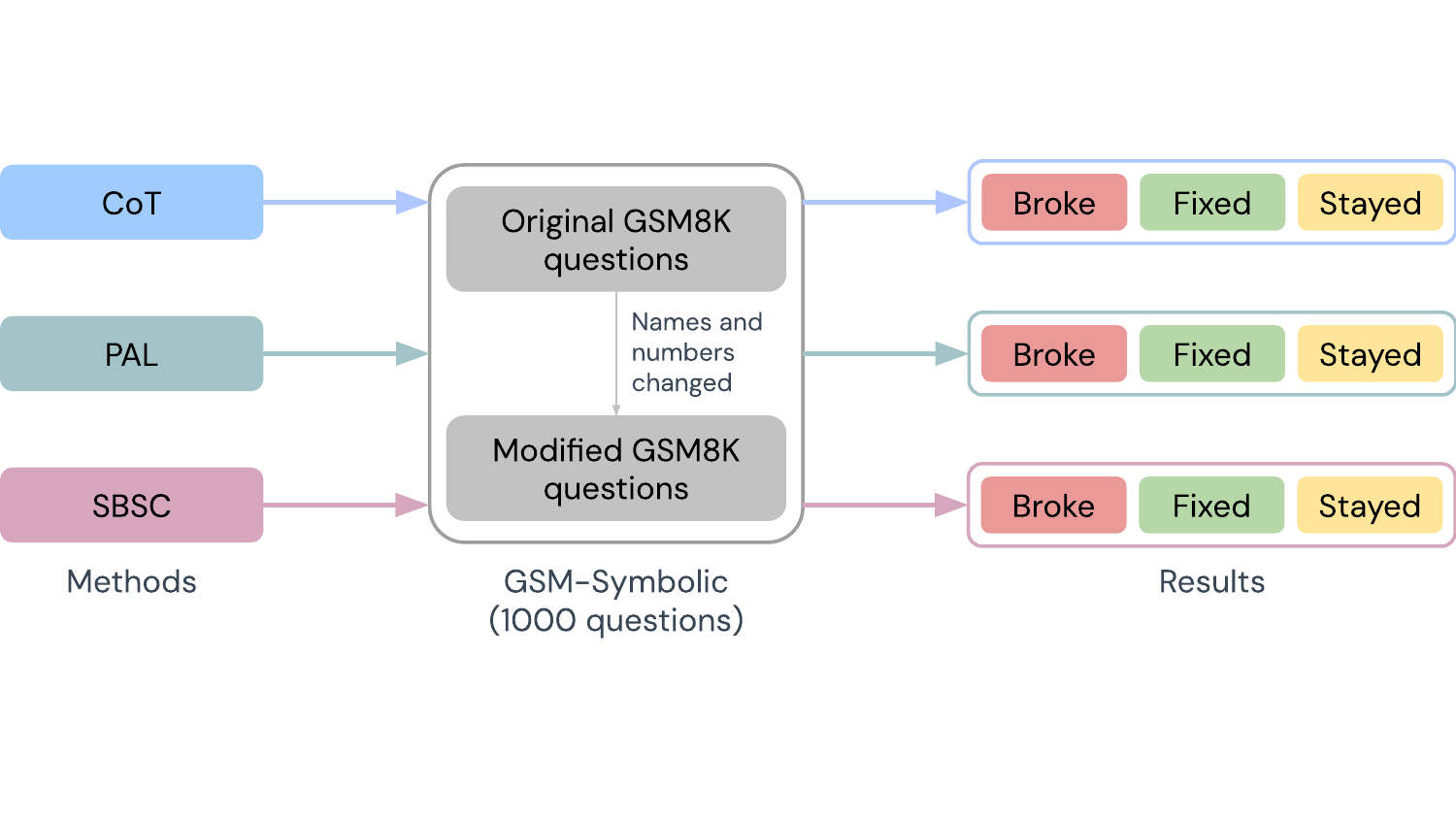}
    \caption{Experimental design overview. Each method was evaluated on 1,000 paired original and modified problems from the GSM-Symbolic main subset. Outcomes were categorized as broke (correct on original, incorrect on modified), fixed (incorrect on original, correct on modified), or stayed (same result on both).}
    \label{fig:method}
\end{figure*}

\section{Literature Review}

In this paper, datasets are fixed sets of problems and their corresponding answers. Benchmarks are the way to evaluate LLMs using datasets, by feeding the model with questions, extracting numerical answers from the output and comparing them to the expected answers. Models are then evaluated on the number of correctly answered questions from the given dataset \cite{cobbe2021,mirzadeh2025}.

Mathematical reasoning benchmarks are crucial for assessing the cognitive capabilities of LLMs. The GSM8K is a dataset of 8,500 grade school math problems that ``[a] bright middle school student should be able to solve'' \cite{cobbe2021}, and it is used to benchmark LLMs on their ability in mathematical reasoning. However, the fundamental flaw in this approach is that when new models are created, they are trained on existing internet data, which might include this dataset.

GSM-Symbolic is another dataset created based on the original GSM8K, with bits of information, such as numbers or names of the subjects in problems, changed. GSM-Symbolics addresses this evaluation flaw by creating problems that have never been on the internet, making it impossible for them to be in training data \cite{mirzadeh2025}. GSM-Symbolic introduces the concept of reasoning robustness, which refers to the ability of LLMs to maintain the accuracy of their mathematical reasoning when facing minor variations in problem formulation that preserve their structure and difficulty \cite{mirzadeh2025}. Large language models exhibit differences in performance of up to 15\% between the original dataset and GSM-Symbolic, with the only difference between the problems in the sets being names and numbers, which do not fundamentally alter the reasoning required to solve them \cite{mirzadeh2025}. More critically, when seemingly relevant but logically irrelevant clauses were added to the problems, performance dropped significantly, by up to 65\% across all state-of-the-art models \cite{mirzadeh2025}, raising concerns about the reliability of models in real-world scenarios and high-stakes applications. The drop in performance suggests that LLMs may rely on pattern matching rather than genuine logical reasoning \cite{mirzadeh2025}.

\IEEEpubidadjcol

As a solution for the identified issue, the code execution approach has emerged. It enables LLMs to generate code that is then executed, and the resulting output is incorporated into the model's reasoning. Program-Aided Language (PAL) models demonstrated that offloading calculations to a Python interpreter while maintaining LLM-based problem decomposition significantly improves accuracy compared to pure reasoning or a chain of thought (CoT) \cite{wei2023,gao2023}.

Iterative code execution is an improvement on this solution. It enables LLMs to write and run code, obtaining the results from its execution. However, in contrast to the single-shot code execution, it then repeats this process multiple times, allowing AI to correct any errors made previously. For example, one implementation of iterative code execution is Step-by-Step Coding (SBSC), which has shown improvements over single-shot code execution on math competition problems \cite{singh2025}.

Mirzadeh et al. \cite{mirzadeh2025} demonstrate through GSM-Symbolic that state-of-the-art models experience a performance drop of up to 15\% when simple changes are made to the problems' names and numbers, while retaining their original structure. Shalyt et al. \cite{shalyt2025} extend this finding to symbolic university-level mathematics with ASyMOB, showing that most models average a 50.4\% performance drop under perturbation. These findings suggest a fundamental limitation in LLM architecture: LLMs perform sophisticated pattern recognition rather than formal reasoning \cite{mirzadeh2025}.

Responding to these limitations, researchers explored code execution to offload calculations to the reliable interpreters, showing mixed robustness results. To better evaluate arithmetic robustness, Gao et al. \cite{gao2023} developed GSM-HARD, a version of GSM8K with larger numbers, which challenges the arithmetic capabilities of the models while maintaining the structure of the problems. They developed and demonstrated how Program-Aided Language models utilize code execution in their thought process, outperforming pure CoT models, with only a 14.3\% drop in accuracy compared to approximately a 70\% drop for CoT models. However, Li et al. \cite{li2024} revealed critical limitations by introducing a GSM-Plus benchmark, which is a variation of the GSM8K dataset that tests mathematical reasoning through eight distinct perturbation types. The results reveal that while program-based models showed only an 11.8\% performance drop on numerical perturbation, compared to 14.0\% for CoT, they exhibited significantly higher vulnerability to the insertion of unnecessary information, ranging from 40.6\% to 43.9\%, compared to 31.8\% to 39.4\% for CoT. The authors conclude that while code execution models excel when arithmetic changes are applied to the problems, they struggle with semantic understanding, challenging the assumption that code execution is a universal tool for improving the robustness of LLMs.

Building on these insights, researchers explored iterative code execution with a feedback approach. Singh et al. \cite{singh2025} introduce Step-by-Step Coding (SBSC), an iterative code execution model with a feedback loop. It showed significant improvements over single-shot code execution and pure reasoning, showing 10.7\% improvement on AMC and 8\% on AIME, one of the highest-level math competitions in the USA. Although this paper did not test the robustness of the approach, it clearly demonstrated the advantage of it over single-shot execution in mathematical problem solving, suggesting its potential improvement of LLM robustness.

\section{Gap in the Research}

Despite significant advances in code-based mathematical reasoning \cite{gao2023,singh2025,li2024}, a critical gap remains in understanding whether these approaches enhance reasoning robustness or merely improve computational accuracy. Although GSM-Symbolic has exposed the fragility of LLM reasoning across problem variations \cite{mirzadeh2025}, existing code execution studies, such as PAL, SBSC, and Tool-Integrated Reasoning (TIR), have evaluated the performance of both iterative and single-shot code execution methods on static datasets like GSM8K and MATH \cite{gao2023,singh2025,li2024}. However, no research has been conducted to evaluate the performance of code execution and iterative code execution on problem variations, such as GSM-Symbolic.

\section{Research Question and Hypothesis}
This study addresses the following research question: Does iterative code execution improve reasoning robustness on GSM-Symbolic's problem variations compared to single-shot code generation and pure reasoning approaches?

I hypothesized that single-shot code execution (PAL) would be the most robust to GSM-Symbolic perturbations, followed by pure reasoning (CoT), with iterative code execution (SBSC) being the least robust. This prediction was based on two prior findings. First, Gao et al. \cite{gao2023} showed that PAL substantially outperforms CoT under arithmetic perturbations on GSM-HARD, suggesting that offloading calculations to a Python interpreter improves robustness when problem variations involve numerical changes. Second, although Singh et al. \cite{singh2025} demonstrated that SBSC outperforms single-shot code execution on competition-level mathematics, its iterative structure was expected to amplify rather than correct misformalization errors on grade-school problems, where a single wrong setup propagates through every subsequent step.

\section{Methodology}

To answer the research question, an experiment was conducted. Three methods were implemented: pure reasoning, single-shot code execution, and iterative code execution. Each method is a set of tools and instructions for the Large Language Model, forcing it to solve problems in different ways. These methods are based on the approaches described in three key papers: the pure reasoning method follows the chain-of-thought (CoT) approach from GSM-Symbolic \cite{mirzadeh2025}, single-shot code execution is based on PAL \cite{gao2023}, and iterative code execution is based on SBSC \cite{singh2025}. Those three methods were evaluated on a GSM-Symbolic dataset, on pairs of original and modified problems (see Figure~\ref{fig:method}). Then, the outputs were compared with the correct answers of those problems, and the number of correct and incorrect answers for each, original and modified, problem was calculated for further analysis.

Each request to an LLM can consist of up to three main parts: the system prompt, the few-shot prompt, and the question itself. The system prompt is a set of instructions for the model to follow. The few-shot prompt containing exemplar problems and solutions is used to specify the desired model output format. For the problems used in this study, an 8-shot prompt (a few-shot prompt containing 8 examples) was used, as is standard for GSM8K and GSM-Symbolic evaluations \cite{mirzadeh2025, cobbe2021}. These 8 example problems were taken from the original GSM8K dataset.

\subsection{Prompting}

First, I constructed an initial prompt (the full message sent to the model, combining the 8-shot prompt and, for some methods, a system prompt) for each of the three methods. For the CoT initial prompt, an 8-shot prompt from Mirzadeh et al. \cite{mirzadeh2025} was used without editing. However, after multiple tests with the LLM model, to achieve the most consistent results, the system prompt was changed from the original ``As an expert problem solver, solve step by step the following mathematical questions.'' to ``Follow the format in the examples provided.'' Also, for the same reasons, the prefill (a technique where the beginning of the model's response is pre-written, forcing it to continue from that point rather than choosing its own opening) ``Let's think step by step'' was added to ensure the model thinks before giving an answer. 

For PAL, the full 8-shot prompt was adapted from Appendix J.4 Gao et al. \cite{gao2023}, which provided all 8 exemplars. However, to ensure reliability and simplicity of extracting the answer from the model output, the format was changed so that instead of the model's code simply outputting the final answer, it was wrapped in a function \texttt{def solution():}, a reusable block of code that calculates and passes the answer back to the script, making it easier to capture automatically. The problems in the 8-shot were the same as in CoT, ensuring consistency, and their order was changed to match the CoT 8-shot \cite{wei2023}. Since PAL did not use a system prompt, I did not use it either. However, to ensure consistency, I had to use a prefill of \texttt{def solution():} to make sure the model generated code immediately. 

For iterative code execution, the prompt was adapted from the SBSC \cite{singh2025} paper; however, for consistency with other methods, the original problems were changed from 4-shot AMC and AIME problems (American math competitions, which contain problems of much higher complexity than GSM8K) to 8-shot GSM8K problems in the same order as in other methods. The example solutions for the 8-shot prompt were generated by Claude AI and later checked by me. The format of the solutions remained mostly intact from Singh et al. \cite{singh2025}; however, for consistency with single-shot code execution, all comments (explanatory notes written within the code) were removed from the code, with meaningful variable names (descriptive names for values in the code, such as \texttt{trees\_planted} instead of \texttt{x}) used instead, as in the original PAL paper \cite{gao2023}. Although the SBSC paper does note the improvement that adding comments makes to the AMC and AIME style problems \cite{singh2025}, this was only demonstrated on competitive-level mathematics problems. On GSM-level problems, meaningful variable names carry most of the explanatory load, consistent with PAL's own math prompt design \cite{gao2023}. 

In this method, the model writes code for one step at a time. The script then runs this code and returns the result of its execution back to the model, which uses it to proceed to the next step. The SBSC system prompt was adapted from Singh et al. \cite{singh2025}, including the key error correction instruction: ``If the executed code snippet returns an error, use it to correct the current step's code snippet. DO NOT restart solving from Step 1.'' A prefill of \texttt{Step 1:} was added to ensure the model began solving immediately in the correct format. The stop word (a specific phrase that signals the script to stop requesting more responses) \texttt{\#\#\#END OF CODE} was kept from the original paper, signalling when the model has reached its final answer. The maximum number of turns (back-and-forth exchanges between the model and the script) was set to 15, matching the original configuration \cite{singh2025}.

\subsection{Implementation}

The Python script for running the experiment was written with the assistance of Claude Opus 4.5 through GitHub Copilot. All three methods were run using Claude Haiku 4.5 through the Anthropic API. Claude was chosen mainly for my familiarity and experience with its API and because relatively few papers in this area use it. The Haiku 4.5 model was chosen primarily for its low cost per query, given the large number of API calls. Because Haiku 4.5 is not the latest state-of-the-art system, it is more prone to errors, making the difference between the methods more significant. All API calls used a temperature setting of 0, meaning the model always selects the most likely next word rather than sampling randomly, ensuring that the results are deterministic and reproducible.

Before the full run, the script was tested multiple times on smaller subsets of problems. Each round of testing revealed new issues in the code execution, answer extraction, and data saving logic, which were fixed iteratively until the script ran reliably. 

\subsection{Data Collection}

The evaluation was conducted on the GSM-Symbolic dataset \cite{mirzadeh2025}, using the \texttt{main} subset, which contains variations where names and numerical values are changed from the original GSM8K problems while the structure and difficulty remain the same. One thousand data points were randomly sampled from the dataset. For each datapoint, the script ran all three methods on both the original problem and its symbolic variation, recording the predicted answer and whether it matched the correct answer, alongside additional information about the problem. Due to an error caught during the initial execution, data for different methods was collected across two separate runs; however, the same 1,000 indices were used across all methods to ensure consistency. All code and prompts are available on GitHub\footnote{\url{https://github.com/masamodelkin/llm-robustness-code-execution}}.

\subsection{Analysis}

To compare robustness across conditions, rather than relying solely on accuracy drops between original and modified problems, each problem was categorized based on how its correctness changed, because aggregate accuracy drops can obscure directional patterns: a condition that breaks many problems but also fixes many could show the same net drop as one that remains stable. Specifically, a problem was categorized as broke if answered correctly on the original but incorrectly on the modified, fixed if answered incorrectly on the original but correctly on the modified, and stayed if answered correctly on both, or incorrectly on both. A chi-squared test of homogeneity was used to determine whether the distribution of these categories differed significantly across the three methods. The chi-square test was chosen because it compares the distribution of categorical outcomes across multiple independent groups. Although the chi-squared test assumed independence between groups, the same problems were used across all three methods. The Limitations section provides a more thorough discussion of this issue.

\section{Results and Discussion}

Results partially supported the hypothesis: while SBSC showed lower robustness than CoT, as predicted, PAL also underperformed CoT, contradicting the expectation that single-shot code execution would improve robustness over pure reasoning. 

\subsection{Quantitative Results}

CoT demonstrated the smallest accuracy drop at 1.3 percentage points, followed by SBSC at 1.4 and PAL at 1.7 percentage points from the original to the modified problems. All three treatments achieved baseline accuracies above 96\%, with CoT leading at 97.9\%, followed by PAL at 97.2\%, and SBSC at 96.4\% (Table~\ref{tab:accuracy}). Although PAL performed better than SBSC on the baseline original problems, it underperformed SBSC in robustness.

\begin{table}[!t]
\centering
\caption{Accuracy on Original and Modified Problems by Method ($n = 1{,}000$ problems per method; drop in percentage points)}
\label{tab:accuracy}
\begin{tabular}{lccc}
\hline
Method & Original (\%) & Modified (\%) & Drop (pp) \\
\hline
CoT  & 97.9 & 96.6 & 1.3 \\
PAL  & 97.2 & 95.5 & 1.7 \\
SBSC & 96.4 & 95.0 & 1.4 \\
\hline
\end{tabular}
\end{table}

\begin{table}[!t]
\centering
\caption{Robustness Distribution by Method ($n = 1{,}000$; chi-square $p = .096$)}
\label{tab:robustness}
\begin{tabular}{lccc}
\hline
Method & Broke & Fixed & Stayed \\
\hline
CoT  & 18 (1.8\%)  & 5  (0.5\%) & 977 (97.7\%) \\
PAL  & 31 (3.1\%)  & 14 (1.4\%) & 955 (95.5\%) \\
SBSC & 25 (2.5\%)  & 11 (1.1\%) & 964 (96.4\%) \\
\hline
\end{tabular}
\end{table}

CoT broke least often at 1.8\%, PAL broke most often at 3.1\%, and SBSC fell in between at 2.5\%; PAL also fixed the most problems (Table~\ref{tab:robustness}). The chi-square test yielded a p-value of 0.096, which is not statistically significant at the standard alpha level of 0.05. Although the observed differences in robustness cannot be confirmed as statistically significant, the data suggests a directional trend worth investigation.

\subsection{Interpretation of Findings}

These results are consistent with the findings of Li et al. \cite{li2024}, who found that code-based methods were more vulnerable to non-arithmetic perturbations on GSM-Plus, particularly to the insertion of unnecessary information. The GSM-Symbolic ``main'' subset tests similar perturbations, such as changing names and numbers, which are closer to semantic changes than pure arithmetic challenges. This may explain why PAL, which was originally shown to excel on GSM-HARD, where only the size of numbers was increased \cite{gao2023}, performed worst in robustness here: the perturbations in this study are not the kind that code execution was designed to handle. 

A likely explanation is that PAL's original advantage, offloading arithmetic to a Python interpreter, is no longer as relevant. On grade-school-level problems, the model handles arithmetic reliably with CoT alone, eliminating the computational edge that code execution once provided. What remains are the failure modes that Li et al. \cite{li2024} identified: when a model formalizes a problem incorrectly into code, the interpreter executes the wrong solution with certainty, whereas CoT reasoning may still arrive at the correct answer through flexible natural language reasoning. 

SBSC's intermediate position is notable. Although Singh et al. \cite{singh2025} demonstrated significant improvements over single-shot code execution on competition-level problems, those gains do not translate directly into robustness on GSM-Symbolic. SBSC's iterative approach cannot fix a fundamentally wrong problem formalization: once the model sets up the wrong equation, iterating on it only yields the wrong answer with greater confidence.

\subsection{Error Analysis}

Examining specific problems where methods diverged reveals patterns consistent with the broader findings. The original version of Problem 91 (Figure~\ref{fig:problem91}) was solved nearly perfectly by all methods. However, on the modified version, PAL and SBSC failed on 9 out of 11 instances while CoT failed on only 2. This is a clear example of the vulnerability that Li et al. \cite{li2024} identified: when surface features of a problem change, code-based methods are more likely to formalize it incorrectly, while CoT adapts more easily through natural language reasoning.

\begin{figure}[!t]
\centering
\fbox{\begin{minipage}{0.93\columnwidth}\small
At Allan's house, there is twice as much corn as cannolis. He has a total of 40 cannolis in his house. Allan bought 60 more cannolis at the store and 40 fewer corns than the number of cannolis. Find the combined total of the number of corns and cannolis Allan has in the house?
\end{minipage}}
\caption{Problem 91 from the GSM-Symbolic dataset.}
\label{fig:problem91}
\end{figure}

Problem 72 asks how many hours a pen pal spends writing letters each week (Figure~\ref{fig:problem72}). The correct answer is 3, but SBSC answered 6 on all 8 instances of the original version, and answered incorrectly on 3 out of 8 instances of the modified problem, while both CoT and PAL answered correctly every time. The model appears to have doubled the writing time by counting both sending and receiving, and SBSC's iterative approach reinforces this error at every step rather than correcting it. The consistent failure pattern directly supports the argument that iteration cannot fix a wrong problem formalization.

\begin{figure}[!t]
\centering
\fbox{\begin{minipage}{0.93\columnwidth}\small
Mike was a pen pal with 5 people. He stopped being penpals with 2 of them. They each send 2 letters a week that are 5 pages long. He responds in kind. He can write a page every 6 minutes. How many hours does he spend writing a week?
\end{minipage}}
\caption{Problem 72 from the GSM-Symbolic dataset.}
\label{fig:problem72}
\end{figure}

Problem 45 (Figure~\ref{fig:problem45}) shows a case where all three methods failed on every instance, consistently producing approximately 162 instead of the correct answer 342. When every method fails identically, the issue lies not in the reasoning approach but in the model's fundamental understanding of the problem. No amount of code execution or iteration can compensate for a misunderstanding that occurs before any reasoning begins, consistent with the conclusion Mirzadeh et al. \cite{mirzadeh2025} that LLMs rely on pattern recognition rather than genuine mathematical reasoning. Similar patterns were observed across several other problems as well. Future research could verify these cases by testing them on human solvers to determine whether the failures reflect genuine model limitations or issues with the dataset.

\begin{figure}[!t]
\centering
\fbox{\begin{minipage}{0.93\columnwidth}\small
Pat has a flower bed that is 111 feet long. Pat wants to fill her flower bed with plants. Pat's flowers grow 12 inches wide so she needs to leave 1.5 feet between every plant. Pat already owns 17 flowers. Each flowering plant costs \$6 at the store, how much money will Pat spend at the store to fill up her flower bed?
\end{minipage}}
\caption{Problem 45 from the GSM-Symbolic dataset.}
\label{fig:problem45}
\end{figure}

\subsection{Fulfilling the Gap in the Research}

Although previous research has tested single-shot code execution on robustness benchmarks like GSM-HARD and GSM-Plus \cite{gao2023, li2024}, no study had evaluated whether iterative code execution improves robustness under perturbations. This study fills that gap by testing all three approaches on GSM-Symbolic. To answer the research question directly: iterative code execution does not improve reasoning robustness compared to pure reasoning, though it did show slightly better robustness than single-shot code execution. The improvements that 
Singh et al. \cite{singh2025} demonstrated on competition-level mathematics do not translate into greater robustness on problem variations. This suggests that although code execution can improve accuracy on hard arithmetic, it does not address the underlying fragility of LLM reasoning that Mirzadeh et al. \cite{mirzadeh2025} identified.

\subsection{Limitations}

This study has several limitations. The chi-square test of homogeneity assumes independence between groups, but the same 1,000 problems were used across all three methods. This means the results for each method on a given problem are not independent, which may affect the validity of the statistical test. 

Due to budget and time constraints, only the ``main'' subset of GSM-Symbolic was tested. The 'main' subset only contains simple names and numerical perturbations, keeping the problems' complexity the same. The P1, P2, and NoOp subsets contain different types of perturbations, such as added complexity, bigger numbers, and irrelevant clauses. Those subsets may have produced different results. Within the ``main'' subset itself, names and numerical perturbations are not separated, meaning that it is impossible to isolate which type of perturbations caused more robustness failures. Given the previous findings that code execution helps with numerical perturbations but struggles with semantic ones \cite{li2024}, a dataset with those perturbation types labelled can reveal the strengths and weaknesses of both single-shot and iterative code execution. 

Since the study was self-funded, only the Claude Haiku 4.5 model was used. Results may differ from other models, particularly stronger or weaker ones. Due to the budget constraints, only 1,000 out of 5,000 available data points were sampled. A larger sample size may have yielded statistically significant results, given that the data showed consistent directional trends. 

Additionally, since GSM-Symbolic was publicly available before Claude Haiku 4.5's training cutoff of February 2025, the dataset may appear in the model's training data. This could disproportionately benefit CoT, which can rely on pattern-matching to memorize solutions, while code-based methods must still generate working code regardless of memorization. This makes it difficult to determine whether CoT's robustness advantage reflects genuine better reasoning or familiarity with the dataset.

\subsection{Implications}

The findings of this study have several real-world implications for the field of LLM reasoning. For researchers, this suggests that robustness should be tested explicitly when evaluating new reasoning methods. The improvements that PAL and SBSC demonstrated on static benchmarks \cite{gao2023, singh2025} did not translate into robustness gains, indicating that strong performance on a fixed dataset does not guarantee consistent performance when problems are varied. Future work on LLM reasoning methods should include perturbation-based evaluation alongside traditional benchmarks. For developers building LLM-based tools, particularly in education or any domain where users phrase the same question differently, these findings suggest caution when adding code execution to a reasoning pipeline. The rigidity of code, where any error in problem formalization is executed with certainty, may make the system less reliable when facing the natural variation in how people ask questions. More broadly, the robustness problem that Mirzadeh et al. \cite{mirzadeh2025} identified is not solved by code execution, suggesting that improving LLM reasoning robustness may require fundamentally different approaches.

\subsection{Areas for Future Research}

Future research should address the limitations of this study in several ways. Testing on P1, P2, and NoOp subsets of GSM-Symbolic would reveal whether code execution methods behave differently on more complex perturbations, particularly the irrelevant clause insertions that cause up to 65\% accuracy drops in the original study \cite{mirzadeh2025}. Running the same experiment across multiple models would show whether the pattern observed here is consistent across LLMs. Most existing research in this area, including the original GSM-Symbolic and PAL studies, was conducted on GPT models \cite{mirzadeh2025, gao2023}, so testing on a broader range of models would provide a more complete picture. Using a dataset with individually labelled perturbations would allow researchers to pinpoint exactly where code execution helps and where it hurts. A larger sample from the dataset may also yield statistically significant results. To address the possibility of dataset contamination, future studies could generate new symbolic variations that have never been publicly available, ensuring that the model cannot rely on memorized solutions.

Perhaps most interestingly, the problems where all three methods failed identically, such as Problem 45, raise the question of whether the failures reflect genuine model limitations or ambiguities in the problems themselves. Testing these problems on human solvers would help distinguish between the two.

\section{Conclusion}
This study set out to determine whether iterative code execution improves reasoning robustness on GSM-Symbolic compared to single-shot code execution and pure reasoning. Three methods, CoT, PAL, and SBSC, were tested on 1,000 problems from the GSM-Symbolic dataset, comparing their accuracy on both original and modified versions of the same problems. The results showed that CoT was the most robust method, with the smallest accuracy drop and the fewest problems that broke under perturbation. SBSC showed slightly better robustness than PAL but still underperformed CoT. Although these differences were not statistically significant, the directional trend was consistent across all measures. These findings suggest that code execution, whether single-shot or iterative, does not improve reasoning robustness on grade-school-level problems, and that the advantages demonstrated by PAL and SBSC on other benchmarks do not extend to resistance against problem variations. The robustness problem identified by Mirzadeh et al. \cite{mirzadeh2025} remains unsolved by current code execution approaches, and addressing it may require fundamentally different methods.

\bibliography{references}
\end{document}